# Assessing Smart Algorithms for Gait Phases Detection in Lower Limb Prosthesis: A Comprehensive Review

Barath Kumar JK and Aswadh Khumar G S

**Abstract—** Over the past few years, the division of gait phases has emerged as a complex area of research that carries significant importance for various applications in the field of gait technologies. The accurate partitioning of gait phases plays a crucial role in advancing these applications. Researchers have been exploring a range of sensors that can be employed to provide data for algorithms involved in gait phase partitioning. These sensors can be broadly categorized into two types: wearable and non-wearable, each offering unique advantages and capabilities. In our study aimed at examining the current approaches to gait analysis and detection specifically designed for implementation in ambulatory rehabilitation systems, we conducted a comprehensive meta-analysis of existing research studies. Our analysis revealed a diverse range of sensors and sensor combinations that demonstrate the ability to analyze gait patterns in ambulatory settings. These sensor options vary from basic force-based binary switches to more intricate setups incorporating multiple inertial sensors and sophisticated algorithms. The findings highlight the wide spectrum of available technologies and methodologies used in gait analysis for ambulatory applications. To conduct an extensive review, we systematically examined two prominent databases, IEEE and Scopus, with the aim of identifying relevant studies pertaining to gait analysis. The search criteria were limited to 189 papers published between 1999 and 2023. From this pool, we identified and included five papers that specifically focused on various techniques including Thresholding, Quasi-static method, adaptive classifier, and SVM-based approaches. These selected papers provided valuable insights for our review.

## 1. INTRODUCTION

The human gait is a highly intricate and recurring process that necessitates the coordinated interaction of muscles, bones, and the nervous system [99]. Its primary purpose is to support an upright posture and maintain balance under both static and dynamic conditions [100]. The gait cycle is defined as the duration from the initial contact of one foot to the subsequent occurrence of the same event with the same foot. In order to address various clinical objectives, multiple partitioning models have been proposed, each offering a different level of detail.

Among these models, the most commonly adopted one consists of two main phases: stance and swing [3–7]. However, certain specific applications require the consideration of a larger number of phases. For instance, three-phase models have been proposed [106–109], as well as four-phase models [110–116], five-phase models [117–122], six-phase models [123–125], and even more complex models involving additional phases [126–131]. The choice of partitioning model depends on the specific research or clinical goals and the need to address particular aspects of gait analysis.

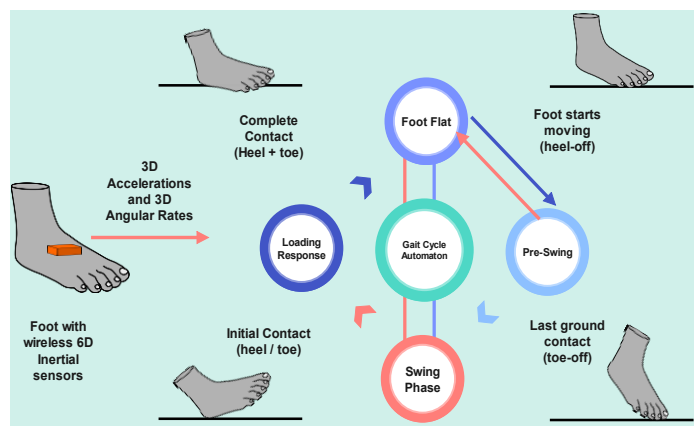

Figure 1. Gait Phase Automaton model

In addition to traditional optical motion analysis systems, researchers have put forward various ambulatory sensor systems that offer alternative approaches to gait analysis. These systems often incorporate specialized technologies such as pressure soles and inertial measurement units (IMUs). These advancements have proven to be especially valuable in the realm of active prostheses and rehabilitation systems, where precise manipulation of robotic forces or the intensity of functional electrical stimulation (FES) needs to be synchronized with the individual's gait

By employing pressure soles and IMUs, these ambulatory sensor systems enable real-time monitoring and analysis of gait parameters shown in figure 2, allowing for dynamic adjustments and interventions during the gait cycle. This capability facilitates the seamless integration of robotic forces or FES interventions, ensuring optimal synchronization with the individual's movement patterns. Such advancements in ambulatory sensor systems have the potential to significantly enhance the effectiveness and functionality of active prostheses and rehabilitation systems, providing more precise and tailored support for individuals undergoing gait-related therapies.

The main focus of this research paper revolves around gait phase detection, specifically employing thresholding techniques. This approach necessitates the measurement of either the ankle joint angle or the foot-to-ground angle. To facilitate this, an Inertial Measurement Unit (IMU) comprising a 3D accelerometer and a 3D gyroscope is assumed to be utilized as depicted in figure 1. Unlike traditional methods like heel switches and pressure soles, the IMU can be conveniently strapped to the body. This feature simplifies its integration into everyday life scenarios, allowing subjects to wear different types of shoes or even no shoes at all.

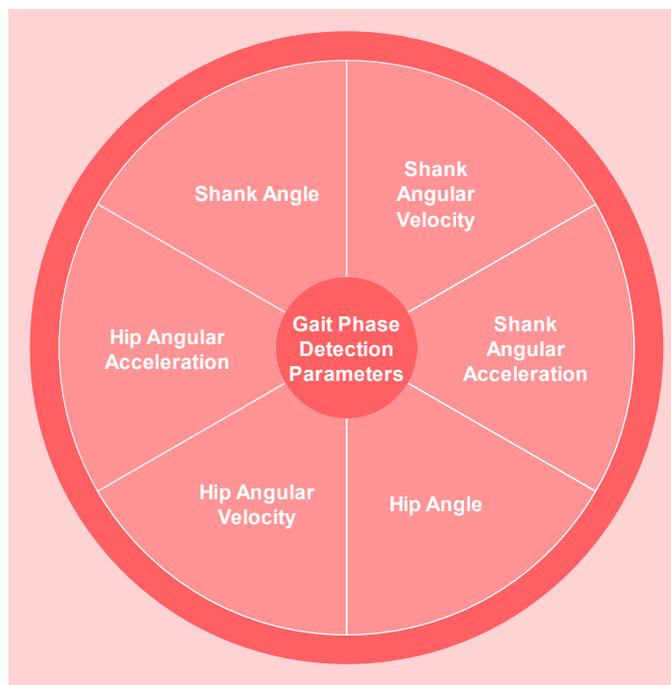

Figure 2. Gait Parameters used for Gait Phase Detection

However, this advantage comes with a trade-off. Due to the variability in sensor placement, the precise orientation of the sensor with respect to the foot cannot be accurately determined. Consequently, there may be some uncertainty regarding the alignment of the sensor and the foot. Despite this limitation, the use of IMUs provides a practical and flexible solution for gait phase detection, making it suitable for real-world applications and enabling individuals to carry out their daily activities without constraints.

## 2. EXPERIMENTAL SECTION

To conduct a comprehensive literature search on the subject of gait phase partitioning, multiple databases were utilized, including Scopus, Google Scholar, and PubMed. The search was performed in June 2023, ensuring a comprehensive coverage of relevant publications up until that point. A range of keywords were employed to facilitate the search process, such as "gait events," "gait phases," and their combinations with terms like "partitioning," "detection," "classification," and "recognition." These carefully selected keywords aimed to capture a broad range of research studies and articles related to the topic of interest. By leveraging these databases and employing specific search terms, the study aimed to gather a substantial body of literature to provide a comprehensive understanding of gait phase partitioning in the field of research at that particular time.

In the process of conducting this systematic review, the articles obtained from the searches were carefully evaluated based on their titles and abstracts. Inclusion criteria were established to ensure that only relevant articles were considered for this study. These criteria included: (i) articles had to be written in English, and (ii) articles had to be published within the timeframe of January 2010 to June 2023 as shown in figure 3. By employing these specific criteria, the review aimed to maintain consistency and focus in the selection process.

It is important to note that the focus of this systematic review was solely on threshold-based algorithms for gait phase detection. Consequently, articles that primarily addressed other methods of gait phase detection were excluded from the study. This selective approach allowed for a more targeted analysis of threshold-based algorithms, providing a deeper understanding of their applications and effectiveness in gait phase partitioning.

To ensure the readability and coherence of the review, a systematic data extraction process was carried out, aligning with the authors' stated intention in the Introduction section. This data extraction involved the identification and categorization of major themes within the selected articles. Three key themes emerged: (i) the granularity of the gait cycle, (ii) sensor placement, and (iii) the method and performance of gait phase classifications.

By categorizing the articles according to these themes, the review was able to provide a focused and organized analysis of the literature. This approach allowed for a comprehensive examination of the different approaches taken in gait phase detection and provided insights into the variations and methodologies used within each theme. By extracting and synthesizing data based on these major themes, the review aimed to present a coherent and informative analysis of the literature, shedding light on the intricacies and advancements in the field of gait phase classification

## 3. OVERVIEW OF IMU BASED GPDs

In contrast to alternative sensor systems, the inertial measurement unit (IMU) offers the additional capability of estimating body segment orientations or joint angles, as demonstrated in previous studies. This capacity becomes particularly valuable in feedback-controlled neuroprostheses. Due to this advantage, the focus in the following discussion is solely on IMU-based methods.

Researchers have developed various techniques utilizing inertial sensors placed on different body regions, such as the trunk, thigh, or shank. Recent advancements have shown promising outcomes, particularly in studies where a shank-mounted IMU was employed. These studies successfully detected three distinct gait phases with high temporal accuracy when compared to a pressure mat system .

By utilizing IMUs in gait analysis, researchers have been able to derive body segment orientations and joint angles, enabling a more comprehensive understanding of human movement as shown in Figure 3.

## 4. SMART ALGORITHMS

In the realm of gait analysis as shown in figure 4, footswitches have emerged as a popular choice for measuring time-related gait parameters. These sensors offer several advantages, including affordability, simple signal conditioning requirements, and high accuracy in gait phase detection. Their widespread use can be attributed to their cost-effectiveness, ease of implementation, and reliable performance in capturing gait phases. In a study by Skelly et al. [109], a real-time algorithm was proposed for detecting five distinct gait phases. The algorithm consisted of two levels. The first level involved a fuzzy logic-based rule set that utilized nine standard gait cycles. This level's primary role was to estimate the five gait phases under consideration. The second level acted as a supervisor, overseeing the duration of the previously estimated gait cycles. If the duration fell below 25% of the mean duration of the standard cycles, they were discarded. The combined approach of fuzzy logic and a supervisor level allowed for accurate gait phase estimation while incorporating a quality control mechanism. By utilizing the rule set based on standard gait cycles and considering the duration of the estimated gait cycles, the algorithm provided a reliable and real-time solution for gait phase detection. Skelly et al.'s study showcased the efficacy of this approach and highlighted the potential of footswitches in enhancing gait analysis through their accurate and efficient measurement of time-related gait parameters.

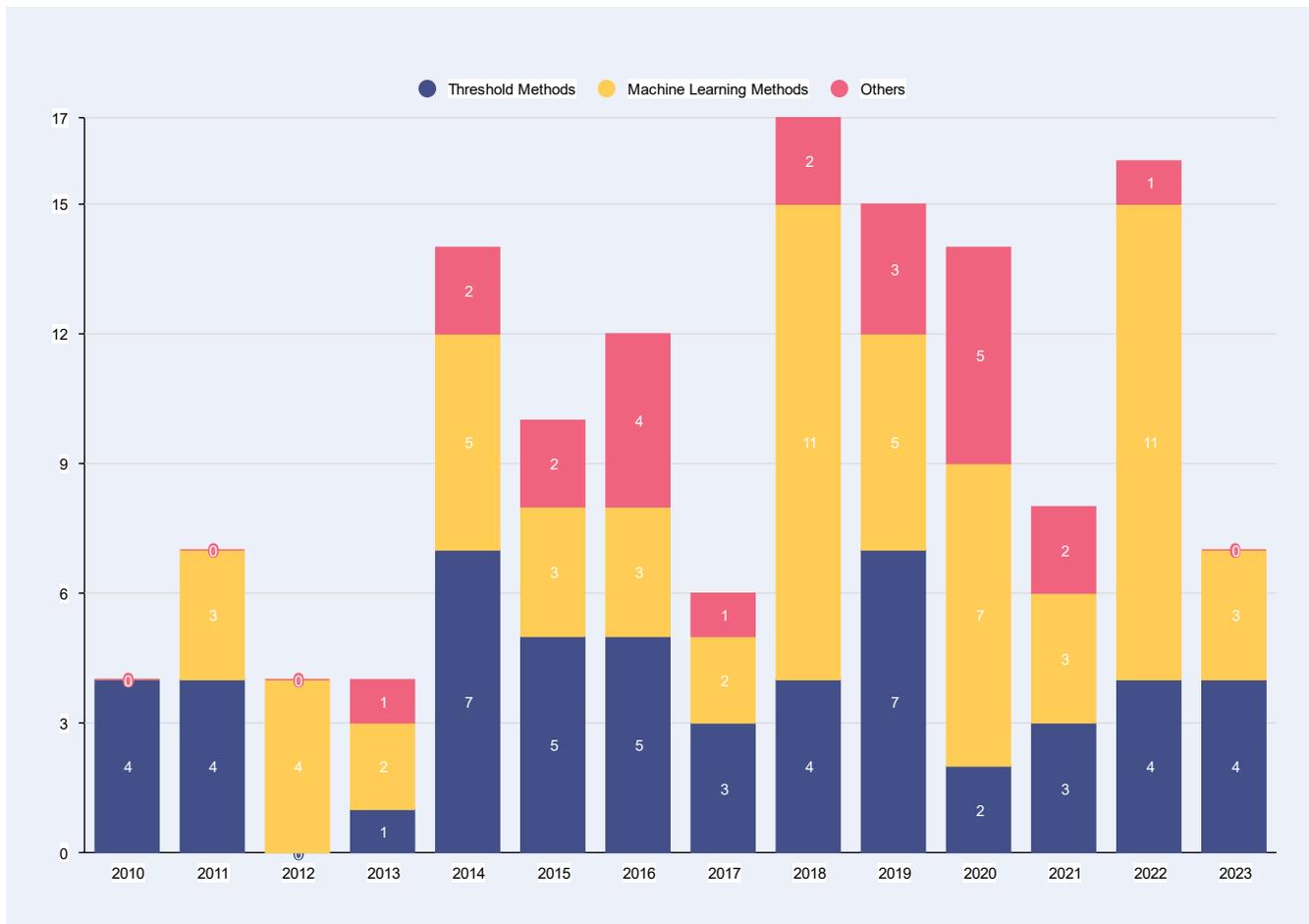

Figure 3. Number of method-based publications from 2010 to 2023

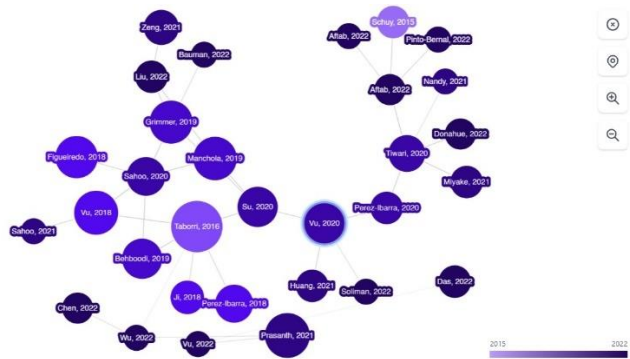

Figure 4. Connection of various papers related to GPD

In their study, Catalfamo et al. [95] proposed a novel method for gait analysis that utilized the activated area in the insole to discriminate between two specific phases as shown in Figure 6. The study involved the testing of two algorithms using walking trials conducted with ten healthy adult participants. The first algorithm, referred to as the force-detection algorithm, aimed to differentiate between the stance phase and the swing phase of the gait cycle. This algorithm employed a weighted threshold value, which was determined by evaluating the maximum and minimum force recorded throughout the entire gait cycle. By comparing the acquired force data to this threshold value, the algorithm determined whether the foot was in the stance phase (when the foot is in contact with the ground) or the swing phase (when the foot is in motion).

The introduction of this method based on the activated area in the insole offered a novel approach to gait analysis. By focusing on the specific phases of the gait cycle and utilizing force data, the algorithm provided a means of accurately discriminating between the stance and swing phases. This advancement could potentially contribute to the development of more precise and effective gait analysis applications, enhancing our understanding of human movement patterns and facilitating the evaluation of gait-related disorders or interventions.

In a study conducted by Selles et al. [93], the researchers demonstrated the feasibility of an algorithm utilizing linear acceleration measured at the shank for the detection of two specific gait phases in exoskeleton control. The study involved the acquisition of longitudinal and antero-posterior linear acceleration data from the shanks of both fifteen healthy adult subjects and ten adult amputees with unilateral transtibial limb loss during level walking. The algorithm developed by Selles et al. relied on identifying key points within the linear acceleration profiles to determine the start and end of the stance phase. Specifically, the first local minimum in the longitudinal acceleration signal corresponded to the initiation of the stance phase, while the last local minimum in the antero-posterior acceleration signal marked the conclusion of the stance phase. To evaluate the algorithm's performance, a subset of participants consisting of five healthy subjects and two amputees was used for training. The remaining participants from both groups were then utilized to test the algorithm. This approach allowed the researchers to assess the algorithm's effectiveness and generalizability across a diverse range of individuals.

In recent years, the research group led by Cappa [106,107,206] has dedicated their efforts to gait partitioning using gyroscope data with the objective of developing a control system for pediatric exoskeletons. Their work has introduced an innovative algorithm that employs a hierarchical weighted decision approach based on the output of multiple scalar Hidden Markov Models (HMMs). This algorithm enables accurate gait phase detection and classification. To evaluate the efficacy of their methodology, Cappa's research group conducted extensive testing on various populations, including healthy adult subjects, healthy children, and children with hemiplegia.

In a study conducted by Lau et al. [103], the researchers assessed the performance of a sensor network comprising various combinations of linear accelerometers and gyroscopes. These sensors were attached to the thigh, shank, and foot regions of the body. The objective of the study was to develop an algorithm capable of determining a three-gait phase model. During the evaluation, the researchers identified specific turning points within the sensor signals. These turning points corresponded to changes in the gradient sign of the signal and proved to be crucial in accurately identifying the required gait events. By applying a threshold method to these turning points, the algorithm effectively distinguished between different gait phases. This study focuses on five different threshold-based algorithms shown in Figure 5.

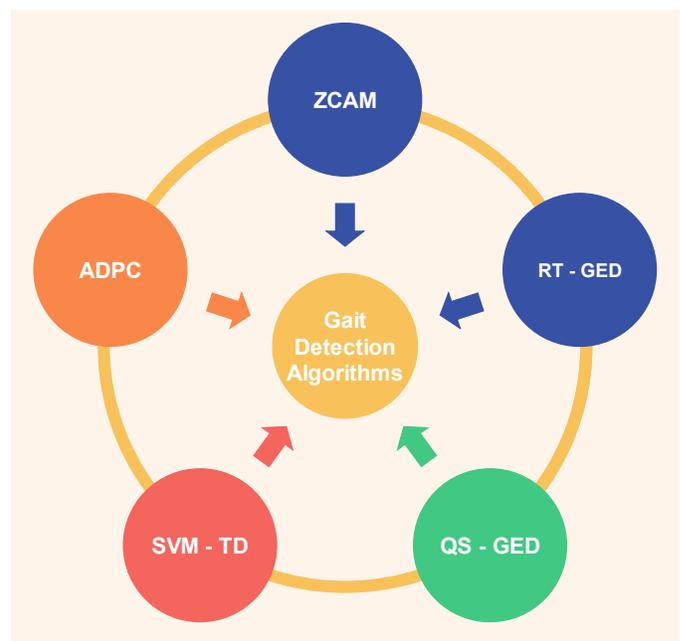

Figure 5. Various Threshold based GPD Methods

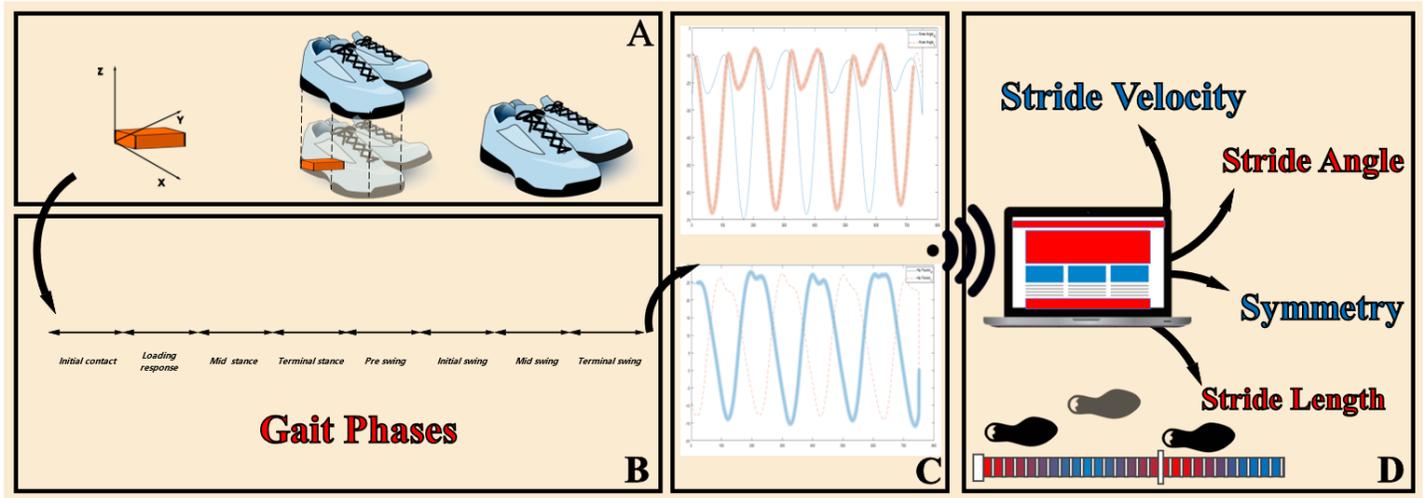

Figure 6. Graphical Illustration of the Data Acquisition System

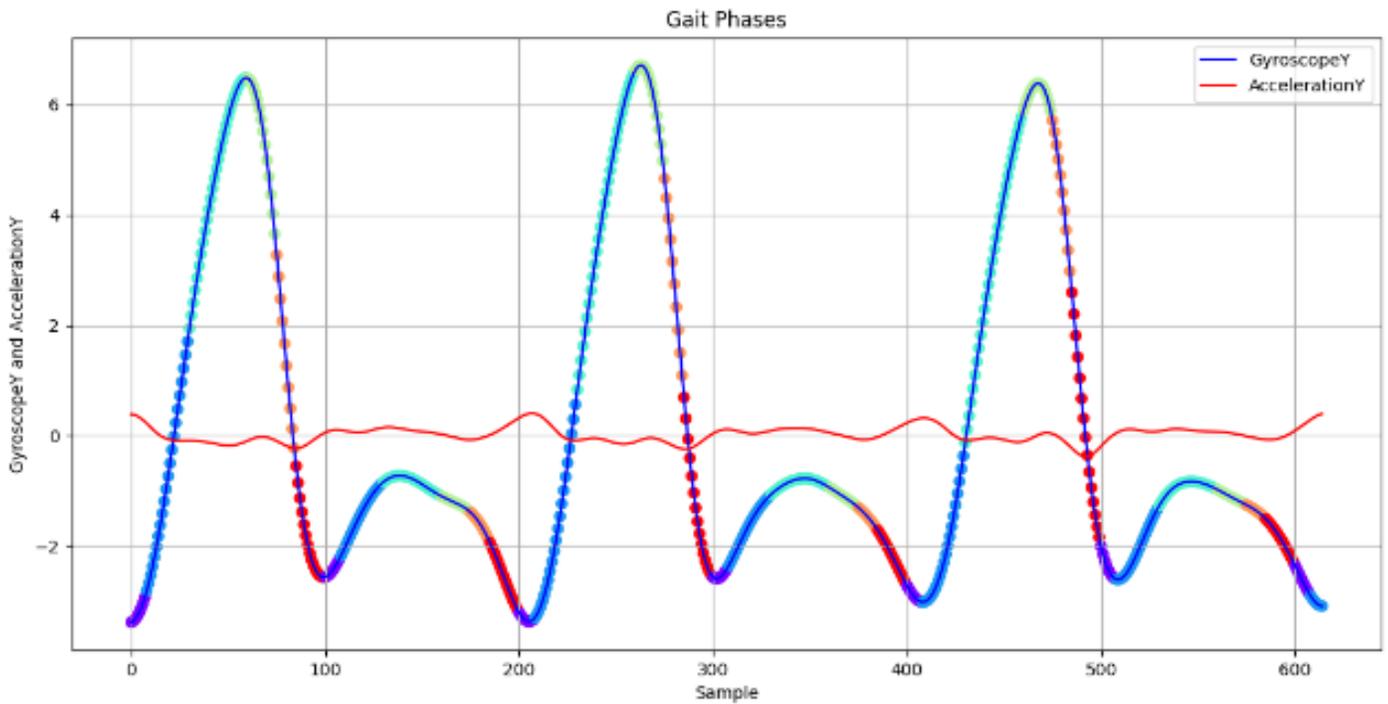

Figure 7. Graphical Phase pattern observed in Right Foot Gyroscope Y

### A. Real Time Gait Event Detection Algorithm

Maqbool et al. developed a heuristic rule-based algorithm using Matlab (R2014a, The Mathworks, MA, USA) for gait event detection. The algorithm focuses on the gyroscope signal from the shank in the sagittal plane, which exhibits a distinctive pattern characterized by two negative peaks on each side of a prominent positive peak. Initial data from two control subjects and a transfemoral amputee were used to develop the algorithm, with a sampling rate of 100 Hz.

To preprocess the data, a second-order Butterworth low-pass filter was applied to the raw signal. The selection of the filter cutoff frequency (fc) took two factors into consideration. Firstly, the fc needed to be low enough to attenuate signal noise, including high-frequency oscillations such as impact spikes during initial contact (IC), in order to minimize erroneous event detection. Secondly, it needed to be higher than the primary frequency of human gait. The proposed algorithm tested cutoff frequencies ranging from 3 Hz to 12 Hz, and 10 Hz was ultimately chosen. This selection aimed to minimize latency in the filtered signal, closely resembling the actual raw signal and reducing any phase shift or delay associated with event detection. The Implementation of this algorithm is shown in Figure 9.

The proposed system holds potential for various applications, including the development of control systems for lower limb prostheses that can switch between control states based on gait phases and events. It could also be employed in outcome evaluation after hip, knee, or ankle replacement surgeries and serve as a diagnostic tool for abnormal and pathological gait analysis during activities of daily living (ADLs). Future work will involve evaluating the algorithm with a larger participant pool, testing it on different terrains and environments such as stair ascent/descent, and assessing its performance during various maneuvers, such as acceleration and deceleration. These advancements aim to enhance the algorithm's effectiveness in functional gait assessment and leverage its outputs for prosthetics control. The accuracy of its detection of gait phases is discussed in Figure 13.

### B. Quasi-Static Detection Algorithm

Gouwanda et al. introduced a wireless sensor network designed for measuring lower extremity motion during walking. The network comprised four wireless Inertia-Link sensors (Microstrain, Inc., Cary, NC, USA) and four transmitters connected to a workstation. Each Inertia-Link sensor was equipped with a triaxial accelerometer and a triaxial gyroscope. For this study, the focus was solely on the gyroscope readings, which had a measuring range of ±5.235 rad/s, bias stability of ±0.00349 rad/s, and a nonlinearity of 0.2%. The onboard microprocessor in each wireless gyroscope performed fundamental data filtering, ensuring minimal jitter in the collected data. The sampling rate was set at 200 Hz, and the transmission range extended up to 10 m.

To facilitate easy mounting, a gait monitoring suit was designed, allowing the attachment of gyroscopes to the lower extremities. The suit featured adjustable straps to ensure a secure and customized fit for each individual, preventing any unwanted sensor movements. The primary objective of the study was to identify gait events based on the angular rate of the shank (ωshank) in the sagittal plane. Hence, the angular rates of the left (ωshankL) and right shanks (ωshankR) were acquired for analysis. The Implementation of this algorithm is shown in Figure 10.

A real-time gait event detection algorithm was proposed, leveraging the gyroscopic data from the human shank. The experimental results demonstrated accurate identification of heel strike (HS) and toe off (TO) events, irrespective of any restrictions imposed on the knee or ankle joints. Future research directions may involve evaluating the proposed algorithm on patients with pathological gait or walking difficulties. Additionally, implementing a predictive mechanism to anticipate HS and TO events could enable the initiation of appropriate functionalities in gait monitoring or assistive devices. The accuracy of its detection of gait phases is discussed in Figure 14.

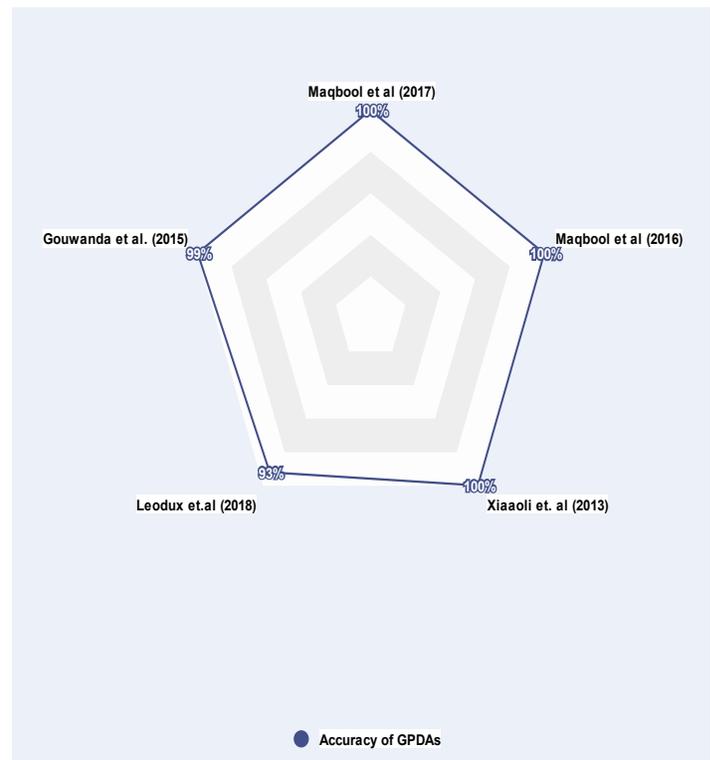

Figure 8. Accuracies of various GPDAs

Fig 10. Implementation of Quasi-Static Algorithm

Figure 9. Implementation of R-GED Algorithm

## C. SVM-Threshold Detection Algorithm

Leodux proposed a robust method for detecting gait events based on observable features in the physical signals. This method relies on signal thresholds to identify specific gait events. Both amputee and healthy gait patterns exhibit common characteristics such as acceleration spikes for heel strike (HS) and toe-off (TO), negative angular velocity during stance, high shank angle at HS, and low shank angle at TO. However, amputee signals may exhibit less pronounced features and a narrower range of shank angles, suggesting the need for different thresholds for each demographic.

The algorithm effectively defines boundaries or "corners" around each gait event using a step-by-step approach. For HS detection, it identifies zero-crossing dips in axial acceleration as an indication of foot contact. Negative angular velocity is then used to confirm the front-to-back direction of leg motion, ensuring safety. Finally, the shank angle must exceed a threshold value, indicating a forward position and confirming HS. Similarly, TO detection relies on a negative spike in axial acceleration followed by an increase above a negative threshold value. Negative angular velocity is again considered to ensure proper forward ambulation, and TO is indicated when the shank angle dips below a negative threshold. The Implementation of this algorithm is shown in Figure 10.

Among the tested algorithms, THR (threshold-based algorithm) demonstrated the highest accuracy for gait event detection and is recommended for use in prosthesis controllers. Its superior accuracy can be attributed to the fact that it targets each gait event specifically for each signal. THR creates separate "corners" or planes for HS and TO in each of the three signals, effectively encapsulating each gait event. The accuracy of its detection of gait phases is discussed in Figure 13.

## D. Adaptive Peak Classifier Method - ADPC

Xiaoli et al. introduced an algorithm that utilizes knee angles and tibia angles in the sagittal plane, estimated by fusing sensor measurements of acceleration, angular velocity, and magnetic field strength. While joint angles can be estimated in multiple planes, only the sagittal plane angles are employed for gait event detection. The algorithm capitalizes on the observation that the foot angular velocity exhibits a maximum peak (tSW) around mid-swing (MSw) in the gait cycle. Preceding this peak, a negative angular velocity peak corresponds to toe-off (TO), while another negative peak occurs at the end of the swing period, indicating heel strike (HS). During heel-off (HO), the angular velocity gradually decreases to a minimum point at TO. Two events that mark the swing phase include foot off (FA), identified when the knee flexion angle reaches its maximum, and toe off (TV), detected when the tibia angle approaches zero in the swing phase.

The algorithm presents a robust approach for detecting seven gait phases during overground walking. Extensive testing on both able-bodied individuals and patients with gait disorders demonstrated the method's reliability. This method can effectively identify abnormalities in the gaits of patients with gait disorders based on the detected gait phases. Furthermore, it serves as a valuable benchmark for evaluating other gait phase detection methods. The Implementation of this algorithm is shown in Figure 11.

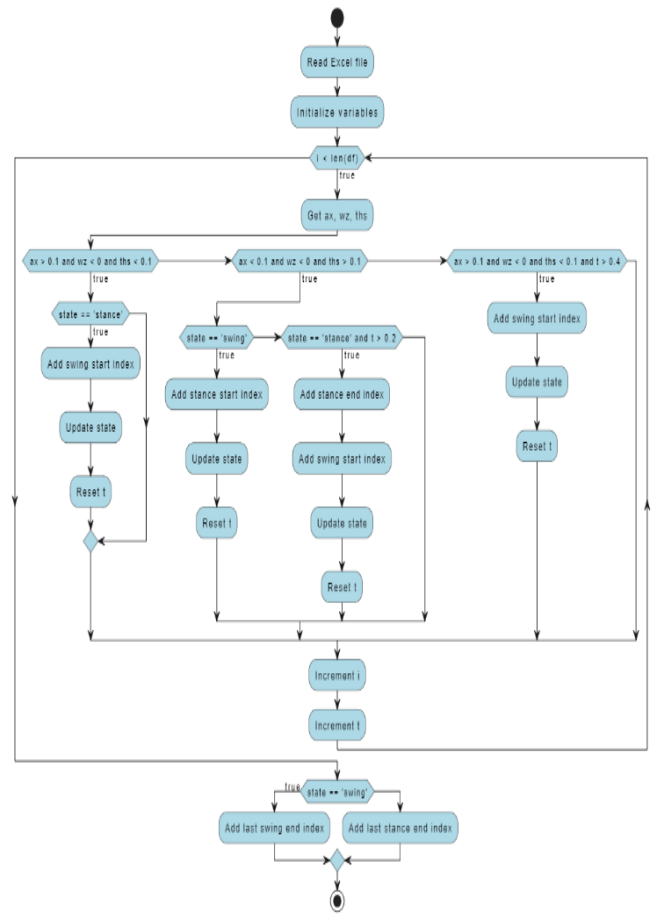

Fig 11. Implementation of SVM threshold Algorithm

In the future, interventions can be directly provided to address identified functional deficits in each gait phase using a two-joint robot (knee and ankle) designed for human use. This approach holds promise for correcting gait abnormalities and enhancing gait performance in individuals with impaired mobility. The proposed algorithm and its associated interventions have the potential to significantly improve the assessment and treatment of gait-related issues.

E. Zero Cross Algorithm Method - ZCAM

Maqbool et al. presented an algorithm that utilizes the gyroscope signal (rotation about the x-axis) and accelerometer signal (acceleration along the z-axis) to identify temporal gait events. Specifically, toe-off (TO) and initial contact (IC) correspond to the two negative peaks observed before and after a maximum peak known as mid-swing (MSW) in the shank's angular velocity signal. The algorithm accurately detects these events.

The shank's angular velocity signal also exhibits a maximum peak during the stance phase, referred to as mid-stance (MST), which occurs when the angular velocity approaches zero. Two additional gait events, foot-flat start (FFS) and heel-off (HO), are identified before and after MST using the acceleration signal. FFS represents the first instance when the foot is flat on the ground during stance. Potential candidates for FFS and HO are determined based on the acceleration signal. During IC, the acceleration signal exhibits peaks and then levels off, while it starts to increase during HO. Concurrently, the shank's angular velocity decreases during dorsiflexion until TO occurs. The Implementation of this algorithm is shown in Figure 12.

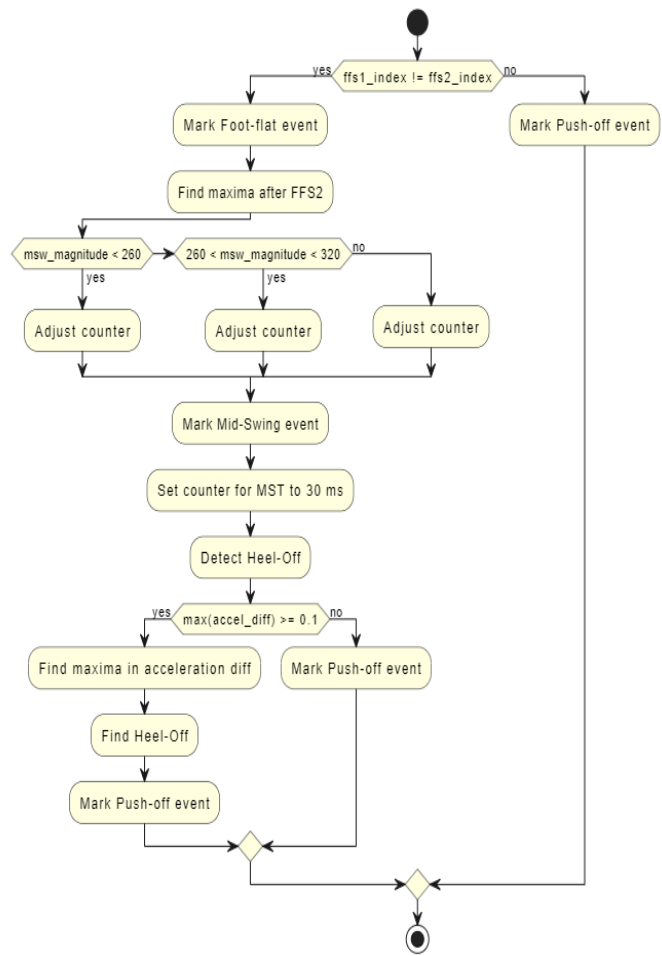

Fig 13. Gait Phase Implementation using ZCAM

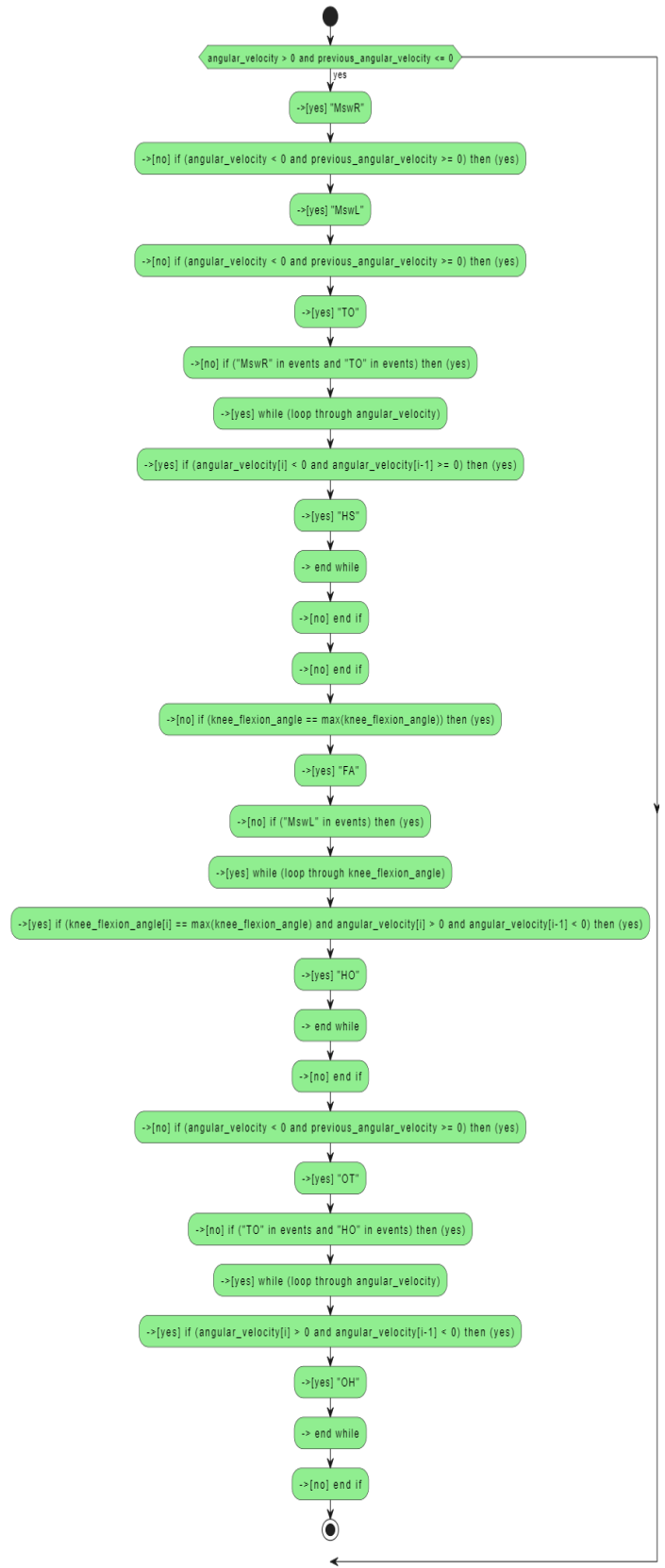

Fig 12. Implementation of ADPC Algorithm

## 5. DISCUSSION

| Authors | Sensor Types | Placements | Methods | Events Detected | Accuracy |
| --- | --- | --- | --- | --- | --- |
| Maqbool et al. [190] (2017) | One IMU, Four Footswitches | Shank and Foot | R-GED | IC, TO, and MSw | 99.78% |
| Gouwanda et al. [70] (2015) | 2 Gyroscopes | Shanks | Quasi-Static | HS, TO | 99% |
| Leodux [191] (2018) | One IMU | Shank | SVM -Threshold | HS and TO | 92.5% |
| Xiaoli et al. [192] (2013) | 4 IMUs | Thigh, Shank, and Feet | Adaptive Peak Classifier | MSw, HS, TO | 100% |
| Maqbool et al. [86] (2016) | One IMU, Four Footswitches | Shank and Foot | Zero Cross Algorithm | IC, TO, and MSw | 100% |

Figure 14. Comparative Analysis of Various Gait Detection Algorithms

Maqbool et al. [190] proposed a method that had a detection accuracy of 99.8% across all gait events. Although the asymmetry behavior could still affect the detection accuracy. The footswitches used in the study were positioned directly underneath the foot, and the participants walked without wearing shoes. However, it is important to acknowledge that the variation in timing differences could also be influenced by the placement of the sensors. It is worth noting that one limitation of the study is the reliance on the prior detection of mid-stance (MSW) before identifying initial contact (IC). This means that IC will not be detected until MSW is successfully identified.

The researchers recognized that the choice of sensor placement can have an impact on the accuracy and timing of gait event detection. Despite using footswitches directly beneath the foot, there may still be variability in the timing differences observed. This variability could be attributed, at least in part, to the specific placement of the sensors on the foot.

In their study, Gouwanda et al. [70] put forward a real-time algorithm for detecting heel strike (HS) and toe-off (TO) events during gait, without imposing any specific restrictions on the knee or ankle. This algorithm holds potential for further research and development in the field. One possible direction for future work is to evaluate the performance of the proposed algorithm on individuals with pathological gait or walking difficulties. This would provide valuable insights into the algorithm's effectiveness in more challenging gait conditions.

Additionally, the implementation of a predictive mechanism could be explored, allowing for the determination of when HS and TO events are likely to occur. Such a mechanism would enable the initiation of appropriate functionalities in gait monitoring or assistive devices, enhancing their overall

Leodux [191] introduced three algorithms for gait event detection, each with a high level of accuracy. Among these algorithms, THR demonstrated the highest overall accuracy, achieving a rate of 92%. Importantly, when trained on three or more healthy subjects or one or more amputees, the THR algorithm did not miss any event detections.

The findings of this study indicate that the THR algorithm is particularly robust and reliable in accurately detecting gait events. It exhibited excellent performance across different subject groups and various walking conditions.

The proposed method by Xiaoli et al.[192] was subjected to a thorough evaluation by an expert. To assess its performance, a data sequence spanning 307 seconds, encompassing 233 gait cycles, was analyzed. Remarkably, the algorithm successfully detected all gait phases without any deletions or insertions, indicating a flawless performance.

The results of this evaluation highlight the robustness and applicability of the proposed method. By successfully detecting all gait phases without any errors, the algorithm demonstrates its potential for practical implementation in various gait analysis and monitoring systems. Future research and validation studies could further investigate the algorithm's performance on larger datasets and across diverse populations, ensuring its reliability and efficacy in clinical and research settings.

Maqbool et al. [86] introduced a method that achieved exceptional results in gait event detection. Their algorithm demonstrated a remarkable 100% accuracy in detecting initial contact (IC) and toe-off (TO) events across five different types of prostheses. This impressive performance underscores the robustness and reliability of the proposed system.

The significance of this achievement lies in its potential application in the development of control systems for lower limb prostheses. By accurately detecting gait phases and events, the algorithm enables seamless transitions between different control states based on the detected events.

## 7. Conclusion

This comprehensive review examined a wide range of studies focused on the current state-of-the-art in gait phase detection. The review provided a valuable synthesis of the existing research and shed light on potential avenues for future investigation. A notable observation was the widespread utilization of inertial measurement unit (IMU) sensors in phase and event detection systems. IMUs were found to be particularly well-suited for long-term applications in daily activities. They offered numerous advantages over electromyography (EMG), including lower energy consumption, increased durability, cost-effectiveness, reduced weight, portability, and ease of placement on the body. Moreover, IMU signals were found to be compatible with all the gait phase detection methods shown in figure 8 and discussed in this review.

The use of gait phase distribution references proved to be valuable for experimental reference and system validation. Notably, the review highlighted that achieving 100% accuracy in detecting two phases was feasible using any of the examined methods. However, when aiming for higher granularity in gait detection, accuracy tended to decrease within the same detection system. To address this challenge, the review suggested the adoption of hybrid algorithms, more complex algorithms, and the incorporation of additional parameters, as these approaches yielded better results.

Signal post-processing emerged as a crucial aspect of gait phase detection to mitigate errors, drift, and signal noise, thereby ensuring high performance. However, it is important to note that introducing additional calculation steps in the post-processing stage could potentially introduce delays, which is a critical consideration for real-time applications where detection latency is a key concern.

Barath et al.: Assessing Smart Algorithms for Gait Phases Detection in Lower Limb Prosthesis: A Comprehensive Review          13Warping and K-Nearest Neighbors Graph Embedding. In Proceedings of the ICASSP 2020–2020 IEEE International Conference on Acoustics, Speech and Signal Processing (ICASSP), Barcelona, Spain, 4–8 May 2020; pp. 1180–1184.

32. Yang, J.; Huang, T.H.; Yu, S.; Yang, X.; Su, H.; Spungen, A.M.; Tsai, C.Y. Machine Learning Based Adaptive Gait Phase Estimation Using Inertial Measurement Sensors. In Proceedings of the 2019 Design of Medical Devices Conference, Minneapolis, MN, USA, 15–18 April 2019.

33. Chakraborty, S.; Nandy, A. An Unsupervised Approach For Gait Phase Detection. In Proceedings of the 2020 4th International Conference on Computational Intelligence and Networks (CINE), Kolkata, India, 27–29 February 2020; pp. 1–5.

34. Lempereur, M.; Rousseau, F.; Rémy-Néris, O.; Pons, C.; Houx, L.; Quellec, G.; Brochard, S. A new deep learning-based method for the detection of gait events in children with gait disorders: Proof-of-concept and concurrent validity. *J. Biomech.* 2020, *98*, 109490.

35. Vu, H.; Gomez, F.; Cherelle, P.; Lefeber, D.; Nowé, A.; Vanderborght, B. ED-FNN: A new deep learning algorithm to detect percentage of the gait cycle for powered prostheses. *Sensors* 2018, *18*, 2389.

36. Gadaleta, M.; Cisotto, G.; Rossi, M.; Rehman, R.Z.U.; Rochester, L.; Del Din, S. Deep Learning Techniques for Improving Digital Gait Segmentation. In Proceedings of the 2019 41st Annual International Conference of the IEEE Engineering in Medicine and Biology Society (EMBC), Berlin, Germany, 23–27 July 2019; pp. 1834–1837.

37. Di Nardo, F.; Morbidoni, C.; Cucchiarelli, A.; Fioretti, S. Recognition of Gait Phases with a Single Knee Electrogoniometer: A Deep Learning Approach. *Electronics* 2020, *9*, 355.

38. Hannink, J.; Kautz, T.; Pasluosta, C.F.; Gaßmann, K.G.; Klucken, J.; Eskofier, B.M. Sensor-based gait parameter extraction with deep convolutional neural networks. *IEEE J. Biomed. Health Inform.* 2016, *21*, 85–93.

39. Su, B.Y.; Wang, J.; Liu, S.Q.; Sheng, M.; Jiang, J.; Xiang, K. A CNN-Based Method for Intent Recognition Using Inertial Measurement Units and Intelligent Lower Limb Prosthesis. *IEEE Trans. Neural Syst. Rehabil. Eng.* 2019, *27*, 1032–1042.

40. Lee, S.S.; Choi, S.T.; Choi, S.I. Classification of gait type based on deep learning using various sensors with smart insole. *Sensors* 2019, *19*, 1757.

41. Agostini, V.; Balestra, G.; Knaflitz, M. Segmentation and classification of gait cycles. *IEEE Trans. Neural Syst. Rehabil. Eng.* 2014, *22*, 946–952.

42. Crea, S.; De Rossi, S.M.; Donati, M.; Reberšek, P.; Novak, D.; Vitiello, N.; Lenzi, T.; Podobnik, J.; Munih, M.; Carrozza, M.C. Development of gait segmentation methods for wearable foot pressure sensors. In Proceedings of the 2012 Annual International Conference of the IEEE Engineering in Medicine and Biology Society, San Diego, CA, USA, 28 August–1 September 2012; pp. 5018–5021.

43. De Rossi, S.M.; Crea, S.; Donati, M.; Reberšek, P.; Novak, D.; Vitiello, N.; Lenzi, T.; Podobnik, J.; Munih, M.; Carrozza, M.C. Gait segmentation using bipedal foot pressure patterns. In Proceedings of the 2012 4th IEEE RAS & EMBS International Conference on Biomedical Robotics and Biomechatronics (BioRob), Rome, Italy, 24–27 June 2012; pp. 361–366.

44. Cherelle, P.; Grosu, V.; Matthys, A.; Vanderborght, B.; Lefeber, D. Design and validation of the ankle mimicking prosthetic (AMP-) foot 2.0. *IEEE Trans. Neural Syst. Rehabil. Eng.* 2014, *22*, 138–148.

45. Cherelle, P.; Junius, K.; Grosu, V.; Cuypers, H.; Vanderborght, B.; Lefeber, D. The amp-foot 2.1: Actuator design, control and experiments with an amputee. *Robotica* 2014, *32*, 1347–1361.

46. Feng, Y.; Wang, Q. Using One Strain Gauge Bridge to Detect Gait Events for a Robotic Prosthesis. *Robotica* 2019, *37*, 1987–1997.

47. Park, J.S.; Lee, C.M.; Koo, S.M.; Kim, C.H. Gait phase detection using force sensing resistors. *IEEE Sens. J.* 2020, *20*, 6516–6523.

48. Jiang, X.; Chu, K.H.; Khoshnam, M.; Menon, C. A wearable gait phase detection system based on force myography techniques. *Sensors* 2018, *18*, 1279.

49. Moulianitis, V.C.; Syrimpeis, V.N.; Aspragathos, N.A.; Panagiotopoulos, E.C. A closed-loop drop-foot correction system with gait event detection from the contralateral lower limb using fuzzy logic. In Proceedings of the 2011 10th International Workshop on Biomedical Engineering, Kos, Greece, 5–7 October 2011; pp. 1–4.

50. Joshi, C.D.; Lahiri, U.; Thakor, N.V. Classification of gait phases from lower limb EMG: Application to exoskeleton orthosis. In Proceedings of the 2013 IEEE Point-of-Care Healthcare Technologies (PHT), Bangalore, India, 16–18 January 2013; pp. 228–23.

Barath et al.: Assessing Smart Algorithms for Gait Phases Detection in Lower Limb Prosthesis: A Comprehensive Review    15